\documentclass[letterpaper, 10 pt, conference]{ieeeconf}  
\usepackage{array}
\usepackage[caption=false,font=normalsize,labelfont=sf,textfont=sf]{subfig}
\usepackage{textcomp}
\usepackage{stfloats}
\usepackage{url}
\usepackage{verbatim}
\usepackage{graphicx}
\usepackage{cite}
\usepackage{xcolor}

\usepackage{cite}
\usepackage{amsmath,amssymb,amsfonts}
\usepackage{algorithmic}
\usepackage{graphicx}
\usepackage{textcomp}
\usepackage{authblk}
\usepackage{gensymb}
\usepackage{hyperref}
\usepackage{scrextend}
\usepackage{changepage}
\usepackage[margin=0.75in]{geometry}
\usepackage{multirow}
\usepackage{amsfonts}
\usepackage{booktabs}
\usepackage{colortbl}
\usepackage{tikz}
\usepackage{pifont}
\usepackage[linesnumbered,ruled,vlined]{algorithm2e}
\usepackage{mathrsfs}
\usepackage[font={footnotesize}]{caption}
\usepackage{makecell}

\IEEEoverridecommandlockouts                              

\title{\LARGE \bf
RaggeDi: Diffusion-based State Estimation of Disordered Rags, Sheets, Towels and Blankets
}

\author{Jikai Ye$^{1,\dagger}$, Wanze Li$^{1,\dagger}$, Shiraz Khan$^{2}$, Gregory S. Chirikjian$^{1,2}$
\thanks{This work is supported by the National Research Foundation, Singapore, under its Medium Sized Centre Programme - Centre for Advanced Robotics Technology Innovation (CARTIN) A-0009428-08-00.} 
\thanks{$\dagger$ Equal contribution}
\thanks{$1$ Department of Mechanical Engineering, National University of Singapore, Singapore}
\thanks{$2$ Department of Mechanical Engineering, University of Delaware, USA}
\thanks{Address all correspondence to G. S. Chirikjian: mpegre@nus.edu.sg, gchirik@udel.edu}}%

\begin{document}

\maketitle
\thispagestyle{empty}
\pagestyle{empty}

\begin{abstract}
Cloth state estimation is an important problem in robotics. 
It is essential for the robot to know the accurate state to manipulate cloth and execute tasks such as robotic dressing, stitching, and covering/uncovering human beings. 
However, estimating cloth state accurately remains challenging due to its high flexibility and self-occlusion. 
This paper proposes a diffusion model-based pipeline that formulates the cloth state estimation as an image generation problem by representing the cloth state as an RGB image that describes the point-wise translation (translation map) between a pre-defined flattened mesh and the deformed mesh in a canonical space. 
Then we train a conditional diffusion-based image generation model to predict the translation map based on an observation. 
Experiments are conducted in both simulation and the real world to validate the performance of our method. 
Results indicate that our method outperforms two recent methods in both accuracy and speed. 
\end{abstract}


\section{INTRODUCTION}
\label{sec: intro}
\textit{State estimation} refers to reconstructing the state of an object 
by processing data collected using sensors. In this paper, we investigate the problem of reconstructing the full state of a \textit{deformable cloth}, such as a rag, using an RGB-D image.
State estimation of deformable (i.e., non-rigid) objects has significant challenges that are not addressed by the conventional state estimation literature.
For instance, the cloth being observed may be folded or crumpled, resulting in large portions of it being occluded in the observed image.
Moreover, a complete description of the cloth requires solving a very high-dimensional state estimation problem, for which the conventional model-based state estimation approaches are intractable. 
The problem of reconstructing the state of a deformable object is an active area of research, since it can facilitate the use of autonomous robotics for services such as folding laundry \cite{abbeelfolding,fabric_vsf_2020,seitabed}, assistive care (e.g., draping a human with a blanket) \cite{BodyUncover,robe}, placing groceries in bags, etc \cite{SURVEY-2022}.
In this paper, we propose \textbf{RaggeDi}, an algorithm for \textbf{Rag} state \textbf{e}stimation by \textbf{Di}ffusion model, which is pronounced as `raggedy'
.

Early work on state estimation of objects was based on \textit{point cloud registration} of rigid bodies, in which a rigid transformation relating a reference shape to the observed point cloud is sought \cite{ICP-1992}.
To apply the point cloud registration method to deformable objects, \cite{CPD-2006} introduced the \textit{Coherent Point Drift (CPD)} method. CPD incorporates the prior knowledge that most deformable objects encountered in the real world have limited elasticity, meaning that the reconstructed point cloud must have a sufficiently smooth (i.e., coherent) structure. The CPD method was further extended to preserve the local structure of the reconstructed point cloud \cite{SPR-2022} as well as to handle partial occlusion of the cloth in the observed image \cite{SPR-Occlusion-2021}. 
Other forms of prior knowledge such as those based on the potential energy of the cloth \cite{EM-2013} and the spatial frequencies of its deformation structure \cite{MODAL-2023} have also been used. 
However, these methods are sensitive to initialization especially when parts of the point cloud are occluded.
Our method provides a reasonable initial guess that can be further refined by these methods.

Another branch of literature has explored the use of deep learning to solve the deformable object state estimation problem. The methods introduced in \cite{NeuralNonRigid2020} and \cite{semiSupervised2020} learn to predict point-to-point correspondences between two RGB-D images of non-rigid objects, whereas the method in \cite{segmentation2020cloth} segments the observed point cloud to learn the corners and edges of the cloth.
In contrast to the aforementioned works, which do not reconstruct the full state of the deformable object, the works in \cite{garmentNet2021,GarmentTracking2023,heldmesh} used the following approach: (i) classify an observed cloth into a pre-defined category (e.g., shirt, pants, etc.),  (ii) complete the point cloud description of the object in the \textit{canonical space},
and finally (iii) map the reconstructed garment back from the canonical space to the observation space. 
Methodologies for closing the sim-to-real gap of mesh-based cloth reconstruction techniques were developed in \cite{ClothReconstruction2023} and \cite{DeformGS2024}, which incorporate dynamics and sequential measurements for iterative refinement of the reconstructed state; however, these methods also rely on a good initial guess of the reconstructed point cloud and/or multiple real-time measurements.
In contrast to these learning-based methods, we propose a new formulation of the cloth state estimation problem as a conditional image generation task so that we are able to utilize the state-of-the-art image generation model, diffusion models.
Since diffusion models scale well with the size of training data, we can easily perform domain randomization to fill the sim-to-real gap. 

In many engineering applications, \textit{diffusion-based generative models} have proven to be an effective way of solving inference problems due to their ability to model complicated prior distributions \cite{song2020score,ho2020denoising}. When the input to the model is augmented with additional information (called as \textit{conditional diffusion} \cite{song2020score}) in the form of measurements, diffusion models can be used to solve state estimation problems where the prior distribution is too complicated to describe using conventional tools. For example, they can learn the prior distribution over a dataset of medical images, which, when combined with sparse and noisy measurements, can be used for fast sample-efficient medical diagnosis \cite{Diffusion-Medical2022}.
Diffusion models are especially well-suited for computer vision tasks \cite{DiffusionVision2023} due to their effectiveness in learning distributions of RGB image frames and have been successfully applied to problems in robotics such as imitation learning \cite{VisuomotorPolicy2023} and motion planning \cite{urain2023se}. 
Our work provides a new application of diffusion models in the context of deformable object state estimation.

The key contributions of this work are: (1) a new formulation of cloth state estimation problem as a conditional image generation task; (2) a diffusion-based generative model to generate the image representing the cloth state based on observation; (3) simulation experiments that demonstrate the effectiveness of the proposed method and real-world experiments that validate its zero-shot sim-to-real transfer capability.

\begin{figure*}[htbp]
\centerline{\includegraphics[width=7in]{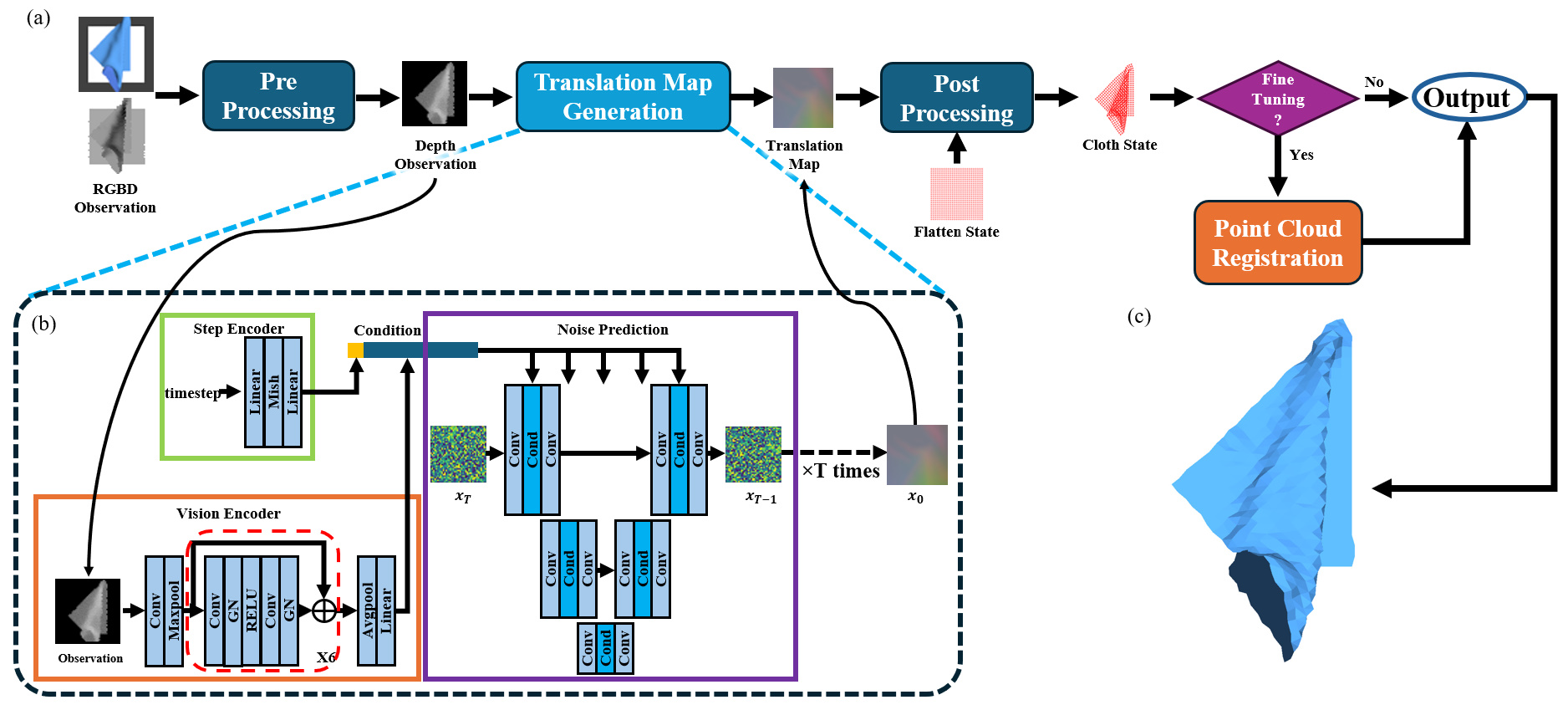}}
\caption{(a) The workflow of the proposed method. (b) The detailed structure of the translation map generation model. It contains three main components: a MLP time step encoder, a ResNet-based observation encoder and a noise prediction network that uses CNN as backbone and the U-Net as the main structure. (c) An example of the mesh predicted by our method. }
\label{fig:pipeline}
\end{figure*}

\section{PRELIMINARY}
\label{sec: preliminary}

In this section, we briefly review non-rigid point cloud registration methods and diffusion models that will be used in the following sections.

\subsection{Non-rigid Point Cloud Registration}
Given two sets of point clouds $X=\{\boldsymbol{x}_1,\boldsymbol{x}_2,...,\boldsymbol{x}_N\}$ and $Y_0=\{\boldsymbol{y}_1, \boldsymbol{y}_2,...,\boldsymbol{y}_M \}$, the goal of non-rigid point cloud registration is to find the translation field $\boldsymbol{v}(\boldsymbol{y}_i)$ (also called velocity function) so that the translated template $Y=\{\boldsymbol{y}_1+\boldsymbol{v}(\boldsymbol{y}_1),\boldsymbol{y}_2+\boldsymbol{v}(\boldsymbol{y}_2),...,\boldsymbol{y}_M+\boldsymbol{v}(\boldsymbol{y}_M)\}$ matches the reference $X$ well.
In CPD \cite{CPD-2006}, the authors model the reference points as a Gaussian-mixture density.
The goal of registration is to maximize the log-likelihood function of $\boldsymbol{v}$ while keeping its smoothness by adding a regularization in Fourier domain
\begin{equation*}
    E({\boldsymbol{v}})\!=\! {-\sum_{n=1}^N \log \sum_{m=1}^M e^{{-\frac{1}{2}\lVert \frac{\boldsymbol{x}_n-\boldsymbol{y}_m-\boldsymbol{v}(\boldsymbol{y}_m)}{\sigma} \rVert}^2}}\!+\frac{\lambda}{2}\!\int \frac{|\tilde{\boldsymbol{v}}(\boldsymbol{s})|^2}{\tilde{G}(\boldsymbol{s})}d\boldsymbol{s}.
\end{equation*}
In structure preserved registration (SPR) \cite{SPR-2022}, the authors add an additional regularization to preserve the local structure of a deformable object.
Both regularizations in CPD and SPR can be written in closed form and the optimization problem can be solved efficiently using the expectation-maximization (EM) algorithm.
However, when some parts of $X$ are unobservable, the registration result usually cannot recover the unobserved state.

\subsection{Diffusion Models}
As a powerful generative method, diffusion models have achieved great success in image generation \cite{ho2022cascaded, gu2022vector} and robot learning \cite{chi2023diffusion, ze20243d}. 
Compared to other deep learning methods, diffusion models demonstrate better training stability and scalability. 
In this paper, we formulate the cloth state prediction as a conditional image generation problem with Denoising Diffusion Probabilistic Models (DDPMs) \cite{ho2020denoising}. 

Diffusion models aim to generate data that obey the distribution $p(\mathbf{x}_0)$ that training data obeys. 
To achieve this, DDPMs start from a Gaussian noise $\mathbf{x}_T$ and perform $T$ denoising iterations with a noise prediction network $\epsilon_\theta$ to recover the output $\mathbf{x}_0 \sim p(x_0)$: 
\begin{equation}
\begin{aligned}
\mathbf{x}_{t-1} = \alpha_t(\mathbf{x}_{t} - \gamma_t\epsilon_\theta(\mathbf{x}_{t}, t)) + z_t
\label{inference}
\end{aligned}
\end{equation}
where $t=T,T-1,...,1$ is the iteration index, $\alpha_t$, $\gamma_t$ are parameters and $z_t$ is a Gaussian noise with zero mean. 
The values of $\alpha_t$, $\gamma_t$ and the covariance of $z_t$ depend on $t$. 

In the training process, firstly an iteration step $t$ is randomly selected, then a Gaussian noise $\epsilon_t$ determined by $t$ is added to a data $\mathbf{x}_0$ from the training set. 
The noise prediction network takes the iteration step and the data sample after adding noise as input. 
The network is trained by minimizing the difference between the output of $\epsilon_\theta$ and the noise: 
\begin{equation}
\begin{aligned}
\mathscr{L} = MSE(\epsilon_t, \epsilon_\theta(\mathbf{x}_{0}+\epsilon_t, t))
\label{MSE_loss}
\end{aligned}
\end{equation}
As proved in \cite{ho2020denoising}, minimizing Eq. \ref{MSE_loss} is equivalent to minimizing the KL-divergence between the actual distribution of the training set and the distribution of data generated using Eq. \ref{inference}. 
More details about diffusion models and DDPM can be found in \cite{ho2020denoising}. 

\section{PROBLEM FORMULATION}
\label{sec: problem_formulation}

We aim to estimate the full state of a folded square-shaped cloth from a depth image.
We assume that we have access to the RGB-D image $(\boldsymbol{o}_{rgb}^{raw}, \boldsymbol{o}_d^{raw})$ of the cloth from a top-down view and that the cloth can be segmented from the background. 
We preprocess the RGB-D image and the resulting depth image $\boldsymbol{o}_d\in \mathbb{R}^{P\times Q \times 1}$ is used as input for our method.
In most cases, only parts of the cloth are observed from this viewpoint due to self-occlusion.

We define the state of cloth as a graph with vertices and edges: $\boldsymbol{s}=(V, E)$.
The vertices are 3D vectors representing the coordinate of points on the cloth in the world frame: $V=\{ \boldsymbol{v}_i \ |\ \boldsymbol{v}_i\in \mathbb{R}^3\}_{i=1:N}$.
The edges are tuples indicating that two vertices are connected, i.e., $(i,j)\in E$ iff $\boldsymbol v_i$ is connected to $\boldsymbol v_j$.
The output of the proposed method is the full mesh of the cloth, which includes both the coordinates of the points and edges; this is in contrast to existing point cloud registration methods that only estimate the former.\footnote{
Note that it is non-trivial to recover the connectivity between points from a point cloud even if it is fully observed \cite{ptstomesh}.} 

\section{METHODS}
\label{sec: methods}

In this section, we present the pipeline of our method illustrated in Fig. \ref{fig:pipeline}. 
In \ref{sec:method_explain}, we explain how we formulate the cloth shape estimation problem as conditional image generation.
The diffusion model structure is detailed in \ref{sec:method_diff}.
Finally, we describe the post-processing that transforms the predicted cloth mesh to the world frame in \ref{sec:post_process}.

\subsection{Cloth Shape Estimation as Image Generation}\label{sec:method_explain}

In this step, we estimate the cloth state in a canonical space, analyzing the overall shape and folding state without considering scale.
The canonical space is a predefined coordinate system where the cloth has a \textit{fixed size} regardless of the true size of the real one, analogous to the imaginary space in the human mind.
The concept has been used extensively in the literature \cite{garmentNet2021,GarmentTracking2023,heldmesh}.

\begin{figure}[htbp]
\centerline{\includegraphics[width=3.2in]{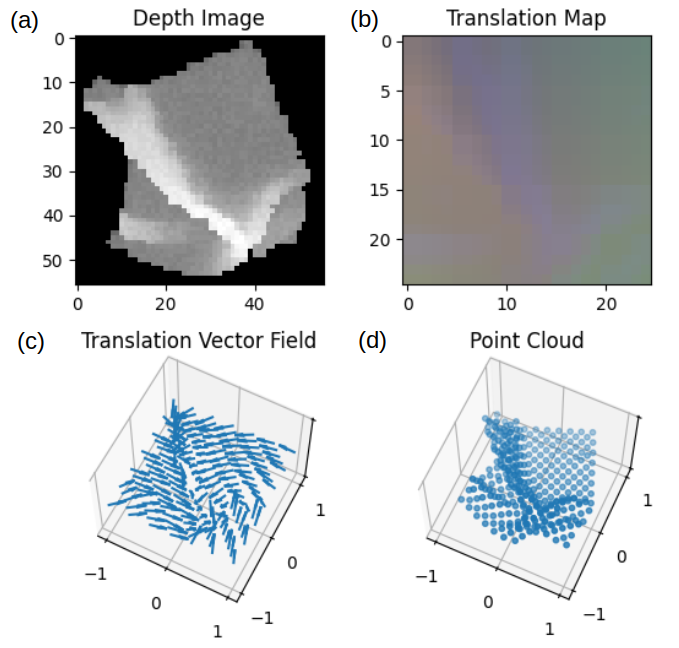}}
\caption{(a) Processed depth image $\boldsymbol{o}_d$ used as the input to the diffusion model. (b) The output translation map $\boldsymbol{\tau}$. (c) A translation vector field that is equivalent to the translation map. (d) Reconstructed predicted point cloud in canonical space using $\boldsymbol{\tau}$.}
\label{fig:method_illustration}
\end{figure}

We reformulate the estimation problem as conditional image generation, where the
key idea is to represent the deformed mesh as an RGB image.
Given a canonical flattened cloth on the $x-y$ plane,
we first discretize it into a fixed $H\times W$ grid.
At the center of each cell, we place a vertex.
Let $\boldsymbol{s}_o^c$ denote this flattened cloth state. 
The flattened cloth can then be represented as an image $im(\boldsymbol{s}_o^c)\in \mathbb{R}^{H\times W \times 3}$ whose pixel value represents the coordinates of the vertex and the adjacency of pixels indicates connectivity.
Any deformed mesh can be generated by translating these vertices to corresponding positions while keeping their connectivity.
We represent the collection of translations as a \textit{translation map} $\boldsymbol{\tau}\in \mathbb{R}^{H\times W \times 3}$, where the pixel value represents the 3D translation of the vertex from the original mesh to the deformed mesh.
As a result, the image representation of the final mesh, $\boldsymbol{s}_f^c$, is calculated by 
\begin{equation}\label{eq:mesh_from_transmap}
im(\boldsymbol{s}_f^c)=im(\boldsymbol{s}_o^c)+\boldsymbol{\tau}\in \mathbb{R}^{H\times W \times 3}.
\end{equation}
To ensure the translation value $\boldsymbol{\tau}(i,j)$ is within a finite range so that we can rescale it back and forth from an RGB value, we keep both the flattened and the deformed mesh centered at the origin of the $x-y$ plane.

We use a physics engine to simulate performing a set of random actions on a cloth.
We collect the RGB-D image and the ground-truth mesh during the process as the data set $D=\{\big((\boldsymbol{o}_{rgb}^{raw})_i,(\boldsymbol{o}_d^{raw})_i, (\boldsymbol{s}_f^c)_i \big)\}_{i=1\sim n}$.
The RGB information is used to segment the cloth from the background in the depth image.
Our diffusion model takes as input the pre-processed depth image $\boldsymbol{o}_d$ and outputs the translation map $\boldsymbol{\tau}$ as illustrated in Fig.\,\ref{fig:method_illustration}.
The final mesh $\boldsymbol{s}^c_f=(V^c,E^c)$ estimated in the canonical space is recovered using Eq. \ref{eq:mesh_from_transmap}.
Note that this mesh estimate is not yet registered with the mesh in the world frame.
We need to perform post-processing to find the correspondence between the canonical vertices and points in the world frame, which is detailed in \ref{sec:post_process}.

\subsection{Translation Map Generation} \label{sec:method_diff}
In this work, the generation of translation map $\boldsymbol{\tau}\in \mathbb{R}^{H\times W \times 3}$ is formulated as a conditional DDPM that is conditioned on the observation $\boldsymbol{o}_d$. 
Therefore, the denoising process of Eq.\ref{inference} becomes: 
\begin{equation}\label{eq:net_inference}
\mathbf{x}_{t-1} = \alpha_t(\mathbf{x}_t - \gamma_t\epsilon_\theta(\mathbf{x}_t, \boldsymbol{o}_d, t)) + z_t
\end{equation}

Fig.\,\ref{fig:pipeline} (b) illustrates the detailed network structure of the diffusion model we use to generate a translation map. 
The pre-processed observation $\boldsymbol{o}_d$ is encoded into a 1D latent vector with the length of 256 by a ResNet-18 \cite{He_2016_CVPR}. 
We replace all the BatchNorm with GroupNorm \cite{wu2018group} as suggested in \cite{chi2023diffusion}. 
The timestep $t$ is also encoded by a MLP encoder with 2 fully connected layers so that the dimension of time has the same order of magnitude with the encoded observation.  
Finally, the outputs of the vision encoder and step encoder are concatenated as a latent vector $\mathbf{c}_t$ to represent the condition in Fig.\ref{fig:pipeline} (b). 

We build the network of noise prediction based on \cite{chi2023diffusion} and use 2D convolutional neural networks (CNNs) as backbone. 
The condition $\mathbf{c}_t$ is passed into each CNN layer of the noise prediction network after further processed by Feature-wise Linear Modulation (FiLM) method \cite{perez2018film}. 
The noise prediction procedure is performed for $T$ times to generate the translation map of the observed cloth.  

\subsection{Post-processing} \label{sec:post_process}
Given the mesh $\boldsymbol{s}_f^c=(V^c, E^c)$ predicted by the diffusion model in the canonical space, we use post-processing to transform it into a mesh in the world frame.
The connectivity does not change during this transformation, since only the coordinates of the vertices are modified.
Suppose we have a calibrated camera, the problem solved during post-processing is equivalent to transforming the predicted vertices from the canonical space into the observed image space.

We first consider the $x-y$ coordinates.
Since the overall shape of the predicted mesh and the observed cloth are similar, the projected mesh should also be similar up to a scale factor.
We project vertices $V^c$ in the canonical space to the $x-y$ plane to form a canonical mask $mask(V^c)$.
Then we renormalize the length and width of the canonical mask $mask(V^c)$ and the observed cloth mask $mask(\boldsymbol{o}_d^{raw})$ by subtracting their means and dividing by standard deviations.
Points in these two rescaled masks have one-to-one correspondence.
These two renormalizing transforms $\mathcal{T}_c$ and $\mathcal{T}_r$ give the transform from the canonical space to the observed image space: $\tilde{\mathcal{T}}_{im}=\mathcal{T}_r^{-1}\circ \mathcal{T}_c$.
Then we transform the canonical mask $mask(V^c)$ to the observed image space $\boldsymbol{o}_d^{raw}$ using $\tilde{T}_{im}$ and perform ICP \cite{ICP-1992} to align this mask with $mask(\boldsymbol{o}_d^{raw})$ in image space for fine-tuning.
We denote the fine-tuning transform as $\mathcal{T}_{ICP}$.
Finally, the transform of the $x-y$ coordinates from the canonical space to the observed image space is 
\begin{equation}
\mathcal{T}_{im}=\mathcal{T}_{ICP}\circ \tilde{\mathcal{T}}_{im}. 
\end{equation}
As for the $z$ coordinate, we calculate an affine transform that matches the maximum and minimum height of vertices in canonical space and the minimum and maximum depth on the observed depth image mask.
For a calibrated camera, we can then transform the vertices in image space back to the world frame using the calibration parameters.

We can further use point cloud registration methods, \textit{e.g.} SPR, to refine the estimated mesh by registering the vertices on the predicted mesh to the point clouds constructed from the depth image $\boldsymbol{o}_d$.
It is observed that the refined mesh is closer to the ground truth than the one without refinement and the one obtained by SPR initializing from a flattened mesh as seen in \ref{sec:sim_exp}.

\section{EXPERIMENTS}
\label{sec: experiments}
This section illustrates the experiments we conduct in both simulation and the real world to validate the performance of the method proposed in this paper. 
Concretely, experiments aim to answer these questions: 
(1) Does the proposed diffusion-based method perform better than our baselines in simulation? 
(2) Can the performance of our diffusion-based model be improved by combining it with a point cloud registration method?
(3) Does the model we train with simulated data transfer in a zero-shot manner to the real-world environment? 
All experiments and data collection are conducted in a computer equipped with an Intel Core i9-14900KF CPU and an NVIDIA RTX 4090 GPU. 
\subsection{Data Collection}
\begin{figure}[htbp]
\centerline{\includegraphics[width=3.5in]{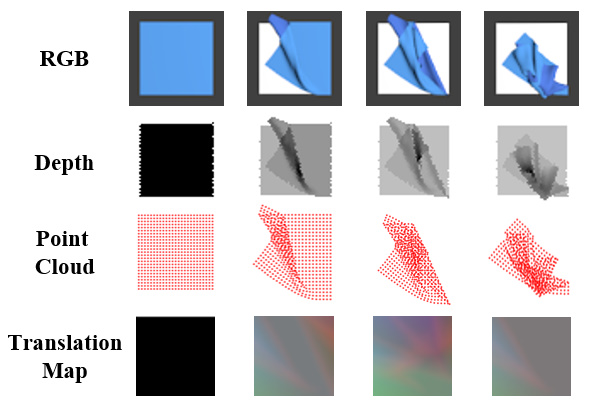}}
\caption{Some examples of data collected in the simulation environment. The first column is the state of the flattened cloth. }
\vspace{-0.3cm}
\label{fig:sim_data}
\end{figure}
In this work, we use the fabric simulator in \cite{fabric_vsf_2020} to collect the data for training and experiments in both simulation and the real world. 
A cloth with the size $1\text{unit}\times1\text{unit}$ is represented as a mass-spring system with a $25 \times 25$ point masses. 
The position of these points describes the full state of the cloth, including the parts that are obscured. 
Meanwhile, the mesh of the cloth at any state can be reconstructed by connecting each point with its 4 nearest neighbors then divided each square face into two triangle faces. 
Since in simulation, the randomization of the cloth state is relatively time-consuming, we collect the data by implementing random pick-and-place actions to the cloth. 
Specifically, we divide the entire data collection procedure into many episodes. 
In each episode, the cloth is initialized as a random state, then a series of random pick-and-place actions are applied to the cloth to generate new states. 
The RGB-D image and the $25 \times 25$ point masses coordinates are recorded after each action. 
We collect the images and point clouds for $123,320$ states within $22,000$ episodes. 
We also collected the state of the flattened cloth (as the canonical mesh $\boldsymbol{s}_o^c$ in Eq. \ref{eq:mesh_from_transmap}) for translation map calculation. 
Some examples of the data collected in the simulation are shown in Fig.\,\ref{fig:sim_data}. 

\begin{figure*}[htbp]
\centerline{\includegraphics[width=7in]{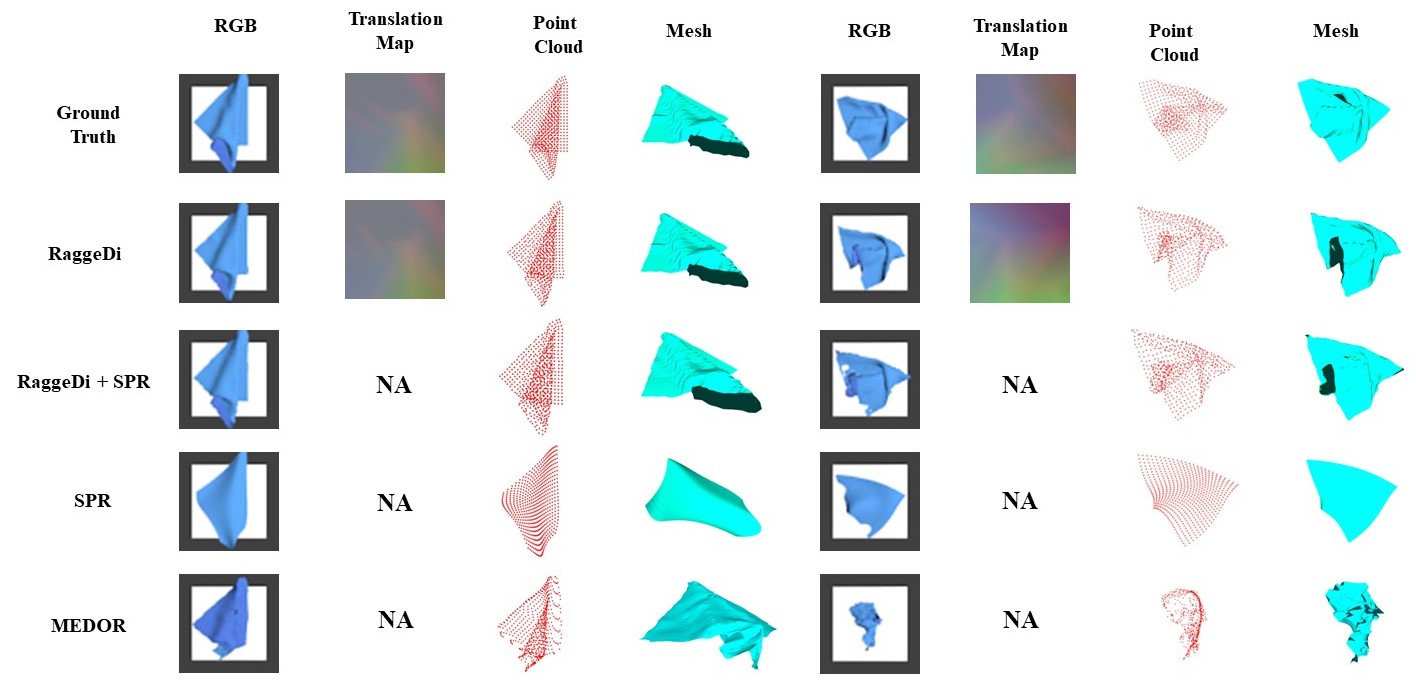}}
\caption{Examples of cloth state predicted with different methods in the simulation experiments. Four columns on the left show the mesh, point cloud, translation map, and RGB images rendered from the mesh of a relatively simple cloth state. Four columns on the right show a complex state. }
\vspace{-0.3cm}
\label{fig:sim_res}
\end{figure*}

\subsection{Data Processing} 

We process the observed RGB-D image taking into account factors that influence the sim-to-real transfer performance, including scale and depth value change.
We first perform a color segmentation that creates a mask for the cloth.
Then, we shift the mask to the center of the depth image and rescale it to fill the whole image.
The value of the background pixels is set to 0.
Finally, we renormalize the cloth pixel values by subtracting their mean, dividing by 3 times the standard deviation, clipping at $[-1,1]$, and rescaling the value to $[55,200]$ for distinguishing points on the cloth from the background.
The resulting normalized depth image is called $\boldsymbol{o}_d$.

For the original flattened cloth mesh $\boldsymbol{s}_o^c=(V^o,E^o)$ and the mesh data $(\boldsymbol{s}_f^c)_i=(V^c_i, E^c_i)$ collected in simulation, we first shift $V^o$ to the origin and rescale the $x$ and $y$ axes to $[-1,1]$ by its width and length $(w,l)$.
Then we shift the center of $V^c_i$ to the origin and rescale the $x$ and $y$ axes with the width and length of $V^o$.
For the $z$ axis, we rescale it using $0.4 w$, which makes the scale of the $z$ axis the same level as $x-y$ but less weighted.
We denote the vertices after rescaling as $\tilde{V}^o$ and $\tilde{V}_i^c$.
The translation map $\boldsymbol{\tau}_i$ is obtained by mapping the translation of vertices $(\tilde{V}_i^c-\tilde{V}^o)$ back to the original grid using $E^o$, clipping the value to $[-3,3]$, and rescaling the value to $[0,255]$.
The resulting depth image and the translation map $D_{train}=\{\big((\boldsymbol{o}_d)_i, \boldsymbol{\tau}_i \big)\}_{i=1\sim n}$ are the training data for our diffusion model.

\subsection{Training Details}
In this work, we use $120,000$ states for model training and the rest are used for testing. 
Only depth images $\boldsymbol{o}_d$ after data pre-processing are used for training. 
Random uniform noises are added to these depth images during training. 
We use DDPM as the noise scheduler, with $100$ timesteps for training and $10$ timesteps at the inference.  
The model is trained for $2,000$ epochs with the batch size $64$, which takes around $20$ hours.
The entire network including input encoder and noise prediction has $6.5$ million parameters. 
The Adam optimizer is used for parameter optimization in training with a learning rate of $1\times10^{-4}$. 

\subsection{Evaluation Metric}
We select the following three metrics for method evaluation: 

\textbf{Chamfer Distance: }The Chamfer distance between two point clouds $S_1, S_2$ is defined as \cite{garmentNet2021}: 
\begin{equation*}
    d_{\mathrm{CD}}\left(S_1, S_2\right)\!=\!\frac{1}{S_1}\! \sum_{x \in S_1} \min _{y \in S_2}\|x-y\|_2^2+\frac{1}{S_2} \!\sum_{y \in S_2} \min _{x \in S_1}\|y-x\|_2^2
\end{equation*}
Smaller Chamfer distance corresponds to higher point cloud similarity. 
We calculate the Chamfer distance between the predicted mesh’s vertices and the ground truth for each method. 
We use the average Chamfer distance over the entire testing set as the final criteria for every method. 

\textbf{Structural Similarity (SSIM): }SSIM $\in [-1, 1]$ is a criterion for evaluating the similarity of two images. 
A higher SSIM corresponds to higher similarity. 
To evaluate the prediction quality of the translation map generation, we calculate the average SSIM value over corresponding translation map pairs predicted by DDPM against ground truth. 

\textbf{Processing Time: }Finally, we compare the average processing time of each state to evaluate the speed of the proposed method.  

\subsection{Simulation Experiments} \label{sec:sim_exp}
\vspace{-0.2cm}
\begin{table}[h]
    \setlength{\tabcolsep}{6pt} 
    \renewcommand{\arraystretch}{1.2}
    \centering
    \caption{Quantitative Results in Simulation}
    \begin{tabular}{cccc}
    \hline
    Methods & SSIM & \makecell[c]{Chamfer\\ Distance} & \makecell[c]{Processing \\ Time (s)} \\
    \hline 
    RaggeDi (ours) & $0.835$ & $\boldsymbol{0.023}$ & $\boldsymbol{0.025}$ \\
    SPR\cite{SPR-2022} & NA & $0.029$ & $2.37$\\
    MEDOR \cite{heldmesh} & NA & $0.045$ & $0.409$\\
    RaggeDi+SPR (ours) & NA & $\boldsymbol{0.019}$ & $2.91$\\
    \hline
    \end{tabular}
    \vspace{-0.2cm}
    \label{tab:sim_res}
\end{table}

The simulated experiment is conducted with $3,320$ random cloth states obtained in the fabric simulator. 
For each state, our method takes the RGB-D image as input, pre-processes the observation, generates the translation map, and recovers the full state. 
We use SPR \cite{SPR-2022} (point cloud registration-based) and MEDOR \cite{heldmesh} (deep learning-based) as our baselines.
For SPR, we register the point cloud of a flattened cloth to the cloth point cloud extracted from the depth image.  
We also combine RaggeDi with SPR by fine-tuning the prediction of RaggeDi using SPR (referred to as RaggeDi+SPR).
For MEDOR, we use the raw depth image as the input and the flattened cloth as the canonical pose.
To keep the training time the same as ours, we train MEDOR using $20,000$ data which takes around $20$ hours in total.
We down-sample the output vertices of MEDOR to $625$ during testing since Chamfer distance is sensitive to the density of points.
We only show the results of MEDOR without fine-tuning because it is observed that fine-tuning does not improve the results much in our task.

The results of the simulated experiment are displayed in Table \ref{tab:sim_res}. 
The average SSIM value of RaggeDi is $0.835$, which demonstrates the high accuracy and stability of our diffusion-based translation map generation model. 
Moreover, RaggeDi outperforms both SPR and MEDOR on Chamfer distance ($0.023$ vs. $0.029$ and $0.023$ vs. $0.045$) at around $95$ times faster than SPR and $16$ times faster than MEDOR. 
The experiment also shows that RaggeDi+SPR (fine-tuning the RaggeDi's output with SPR) can further reduce the Chamfer distance ($0.023 \rightarrow 0.019$). 
Fig.\,\ref{fig:sim_res} illustrates some predicting results of different methods. 
Both RaggeDi and RaggeDi+SPR can predict the points belonging to folded and obscured portions on the cloth. 
However, the results of SPR are smooth and lose information about the obscured parts, and the results of MEDOR do not capture the shape well.

\subsection{Real World Experiments}
\begin{figure}[htbp]
\centerline{\includegraphics[width=3.2in]{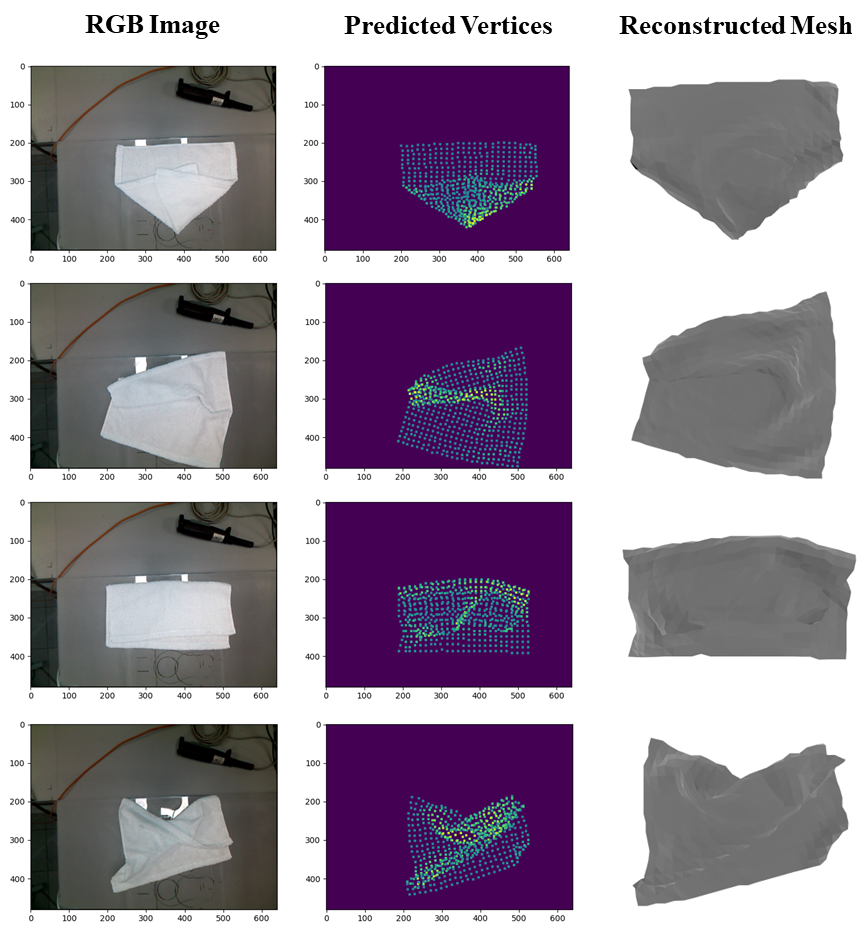}}
\caption{Real-world experiment results are shown above. The left column is the observed RGB image. The middle column is the vertices predicted by RaggeDi. Post-processing transforms the mesh into the image space. The right column shows the visualized mesh.  }
\vspace{-0.2cm}
\label{fig:real}
\end{figure}

We mount a PrimeSense RGB-D camera on a Franka robot arm which moves to a given height and orientation that is parallel to the ground for capturing the RGB-D image.
To make the cloth mask segmentation easier, we choose a white square rag and put it on a transparent box.
Some observed RGB images are shown in Fig.\,\ref{fig:real}.
We perform color and depth segmentation to obtain the rag mask.
We test RaggeDi without SPR refinement in the real world experiment.
Due to the difficulty of obtaining the ground-truth mesh of the rag, we only show qualitative results.

The experiment results are shown in Fig.\,\ref{fig:real}.
The predicted vertices after post-processing capture the overall shape and position of the ground truth cloth.
For the first three tasks which are relatively easier, the reconstructed meshes well capture the folding states.
For the complicated case in the last row, the reconstructed mesh shows the crease on the bottom left and bottom right.
But the folding state on the top right is not well captured.
Note that in some reconstructed meshes, self-penetration is observed.
It could be removed by the method proposed in \cite{SPR-Occlusion-2021}.
To keep the simplicity of the presentation, we do not include that in the post-processing.

\section{CONCLUSIONS}
\label{sec: conclusions}
In this paper, we propose a novel deep learning method ({RaggeDi}) to estimate the cloth state based on DDPM. 
The proposed method takes one RGB-D image as input. 
After pre-processing, the depth image of the cloth is input to the DDPM which generates the translation map between the full state of the cloth and a pre-defined flattened cloth. 
Finally, the state of the cloth is recovered from the predicted translation map. 
Experimental results in both simulation and the real world show that our approach outperforms the baselines in both accuracy and speed.  
Moreover, the estimation performance can be further increased by combining RaggeDi with a point cloud registration method. 
Future work will focus on exploring more complex scenarios, such as estimating the clothes with different topologies and occluded by other objects using diffusion models on meshes \cite{liu2023meshdiffusion}.





\bibliographystyle{IEEEtran}
\bibliography{reference.bib}



\end{document}